%% file: paper.tex
\documentclass[11pt]{article}

\usepackage[final]{acl}

\usepackage{times}
\usepackage{latexsym}
\usepackage[T1]{fontenc}
\usepackage[utf8]{inputenc}
\usepackage{microtype}
\usepackage{inconsolata}
\usepackage{graphicx}
\usepackage{subcaption}
\usepackage{booktabs}
\usepackage{array}
\usepackage{tabularx}

\title{One Feedback System Does Not Fit All: Localising Data-to-Text Driver Coaching for the United Kingdom and Nigeria}

\author{
  Iniakpokeikiye Peter Thompson$^{1,*}$ \quad Jawwad Baig$^{1,*}$ \quad
  Ehud Reiter$^1$ \quad Dewei Yi$^1$ \\
  $^1$Department of Computing Science, University of Aberdeen, \\
  King's College, Aberdeen AB24 3UE, United Kingdom \\
  \texttt{iniakpokeikiye.thompson1@abdn.ac.uk} \\
  \texttt{\{j.baig.18,e.reiter,dewei.yi\}@abdn.ac.uk} \\
  $^*$Equal contribution
}

\begin{document}
\maketitle

\begin{abstract}
Data-to-text driver coaching is often presented as a generic pipeline from telematics events to advice. This paper argues that its content requires localisation because usefulness and credibility depend on drivers' knowledge, prevalent risks, regulation, infrastructure, and available data. Two independently developed systems in the United Kingdom and Nigeria are compared by tracing requirements through content selection, generation, and field evaluation. The UK system prioritises post-trip reflection, explanations tied to road and place context, and tone-sensitive wording. The Nigerian system combines legally grounded, once-daily Tips based on detected events with weekly persuasive Reports; it foregrounds safety education and alcohol-related risk in response to reported gaps in formal training and traffic-rule knowledge, as well as local road-safety priorities. Reliable speed-limit metadata supported speeding feedback in the UK, whereas its scarcity led the Nigerian evaluation to exclude speeding from its outcome metric. Both interventions were associated with reduced distance-normalised unsafe-event rates in their own field studies, although their designs and metrics preclude an effect-size comparison. The analysis yields a requirements-to-content design process for localising safety-critical NLG without treating a high-income deployment as the default.
\end{abstract}

\input{sections/introduction}
\input{sections/related_work}
\input{sections/context}
\input{sections/systems}
\input{sections/architecture}
\input{sections/evaluation}
\input{sections/discussion}

\section*{Acknowledgments}

Peter Thompson's doctoral research was funded by the Tertiary Education Trust Fund (TETFund), Nigeria, with institutional support from Niger Delta University, Nigeria.

\bibliography{refs}

\end{document}

%% file: sections/introduction.tex
\section{Introduction}

Natural Language Generation (NLG) has long been used to turn structured data into readable feedback for users who need to act on complex information \citep{reiter2000building,osuji2024datatotext,sharma2024neural}. In driver-feedback systems, this task is safety-critical: the text must be understandable and motivating, but it must also remain grounded in sensor observations, road context, and legal constraints. Reviews of driver-feedback systems and post-trip interventions show that sensing, presentation, timing, and behaviour-change mechanisms must be designed together \citep{riostorres2016overview,michelaraki2021posttrip}. The problem becomes harder when systems are deployed across different infrastructure and regulatory environments.

This paper compares two NLG-based driver-feedback systems described in recent doctoral system studies and related publications: DRIVINGBEACON, a United Kingdom (UK) system for post-trip telematics feedback \citep{sourceStudyA2026,sourcePubA2022}, and Safe Drive Africa, a Nigerian mobile system for culturally attuned safety feedback \citep{sourceStudyB2025,sourcePubB2025}. Both convert sensed behaviour into textual coaching, but they were designed from different local evidence rather than by translating a fixed intervention.

The central claim of this paper is that localising data-to-text coaching is fundamentally a problem of content selection, grounding, and framing rather than language translation. UK participants sought fair, contextual reflection that explained what happened, where, and why, without moralising. Nigerian stakeholders identified gaps in formal training and traffic-rule knowledge alongside alcohol-influenced driving, creating a distinct requirement for foundational education, explicit legal grounding, and alcohol-aware coaching. Architecture and sensor availability enable and bound these content strategies: reliable speed-limit metadata in the UK support speeding narratives, whereas data scarcity in Nigeria requires omitting ungrounded speed claims. Implementation differences alone are not the primary contribution; rather, we synthesize how contextual requirements dictate architectural tradeoffs.

Our contributions are:
\begin{itemize}
    \item a requirements-to-content comparison showing how locally elicited risks, knowledge, and expectations shaped two NLG interventions;
    \item a synthesis of how road-data availability and safety priorities determine responsible content boundaries and evaluation targets;
    \item field evidence of behavioural effectiveness from both deployment contexts; and
    \item a five-stage design process, derived from the two deployments, providing a structured methodology for eliciting requirements, auditing evidence, defining messages, enriching content, and generating reports when localising safety-critical NLG systems.
\end{itemize}

%% file: sections/related_work.tex
\section{Related Work}

Early NLG systems for behaviour change used controlled templates and explicit tailoring to preserve traceability \citep{reiter2003smoking,williams2008skillsum}. Driver feedback followed a similar path: early data-to-text work generated explanations directly from telematics before SaferDrive integrated controlled content selection into a behaviour-change intervention \citep{braun2015creating,braun2018saferdrive}. Reviews now distinguish real-time warnings from post-trip reflection and show wide variation in sensing, feedback modality, personalisation, and evaluation design \citep{riostorres2016overview,michelaraki2021posttrip,boylan2024telematics,kontaxi2025feedback}. Field studies nevertheless provide convergent evidence that telematics feedback can change measured behaviour, including randomised financial-incentive feedback, personalised programmes for young drivers, and safe-driving apps \citep{stevenson2021telematics,meuleners2023personalized,vankov2023provisional,warren2018backpocketdriver}.

Beyond research prototypes, usage-based insurance (UBI) systems turn telematics into scores, incentives, and post-trip advice. Surveys of telematics research cover risk modelling, safer transport, and insurance applications \citep{gao2022telematics,ghaffarpasand2022vehicle}. User acceptance depends not only on usefulness but also on trust, social influence, and privacy \citep{quintero2023ubi,picco2023monitoring}. Automotive-data studies accordingly emphasise user control, data minimisation, and governance when monitoring can affect premiums or other consequential decisions \citep{dowthwaite2024privacy,truby2024regulatory}. Recent systems have generated actionable risk-mitigation suggestions from insurer data \citep{li2023driving}, while a large randomised field experiment found that weekly UBI-style feedback and incentives reduced several measured risky behaviours \citep{ebert2026ubi}.

Recent Large Language Models (LLMs) make it easier to produce fluent and adaptive feedback, but they introduce risks when outputs must be factually defensible. Evaluation in high-stakes summarisation and feedback domains therefore encompasses factual consistency, task suitability, and human interpretation rather than only surface fluency \citep{novikova2017need,vanderlee2019best,croxford2025llm}. Hybrid architectures are a practical response: structured components decide what can be said, while neural components may adapt how it is said.

Persuasive technology research also matters because feedback is intended to influence future behaviour, not merely describe past events \citep{fogg2009behavior}. Meta-analytic road-safety evidence supports the relevance of attitudes, norms, and perceived behavioural control, while person-based intervention design argues for iterative grounding in target users' needs and contexts \citep{somoray2024tpb,yardley2015person}. Safe Drive Africa explicitly draws on the Theory of Planned Behaviour (TPB), using prompts that address attitudes, norms, motivation, and perceived behavioural control \citep{ajzen1991theory,sourceStudyB2025}. DRIVINGBEACON instead uses predefined user attributes to adapt tone while keeping factual claims under rule-based control \citep{sourceStudyA2026}.

Most work asks whether feedback is tailored to an individual. The present comparison adds a different level: the suitability of the \emph{content strategy} for its socio-technical setting. Persuasive strategies are not received uniformly across cultures, as comparisons of Nigerian and Canadian users demonstrate, and broader driving research links behaviour to culture, income, and governance \citep{oyibo2018persuasive,arslan2024culture}. Recent cultural-NLP research likewise distinguishes cultural adaptation from literal translation and warns that culture comprises multiple, context-dependent elements requiring explicit evaluation \citep{singh2024cultural,liu2025cultural}. Localisation here therefore extends beyond cosmetic translation or the addition of place names. It can change which behaviours are prioritised, whether advice provides rule education or supports reflection, which evidence establishes credibility, and which outcomes can be measured reliably. The two systems are examined as an analytic case comparison rather than as a controlled cross-country experiment.

%% file: sections/context.tex
\section{From Local Context to Content}

Table~\ref{tab:context} summarises the evidence that motivated the two content strategies. The reported patterns describe the respective requirements samples and deployments; they do not characterise all UK or Nigerian drivers.

\begin{table*}[t]
\small
\centering
\begin{tabularx}{\textwidth}{@{}p{0.16\textwidth}XXp{0.25\textwidth}@{}}
\toprule
Design question & UK evidence & Nigerian evidence & Consequence for generated content \\
\midrule
What help is needed? & Drivers sought fair, actionable explanations of detected events and preferred post-trip reflection to distracting live alerts. & Stakeholders, specifically drivers and Federal Road Safety Corps (FRSC) officials, identified lack of training and knowledge of traffic rules as causes of unsafe driving, alongside demand for education and awareness. & UK reports explain behaviour in context; Nigerian Tips explicitly teach rules and consequences. \\
Which risks deserve emphasis? & Speeding, phone use, harsh manoeuvres, and risk near schools and other busy or sensitive places. & Non-compliant/aggressive driving, distraction, fatigue, and alcohol/substance use; the app could reliably instrument harsh braking, acceleration, swerving, and alcohol risk. & Each system selects locally relevant, credibly detectable behaviours rather than a universal inventory. \\
What makes feedback acceptable? & Supportive, non-judgemental language; clarity, fairness, and sensitivity to preferred feedback style. & Credible Nigerian legal references plus respectful, ability-building coaching that acknowledges local road realities. & Personalised tone in the UK; two complementary Nigerian surfaces for authority and persuasion. \\
What can data support? & Telematics supplied speed and speed-limit fields plus mapped/geofenced context. & Digital posted speed limits were too sparse and unreliable for consistent legal-speed classification. & UK content can discuss speeding and place risk; the Nigerian event rate and feedback avoid unsupported speeding claims. \\
\bottomrule
\end{tabularx}
\caption{Local evidence and its consequences for content. The contrast concerns study populations and deployment evidence, not national stereotypes.}
\label{tab:context}
\end{table*}

\subsection{UK: reflection in a formal driving environment}

Complementary methods addressed distinct requirements before implementation. Two structured surveys elicited report-format preferences and perceived influences on driving from the same 30 active UK drivers, recruited through personal and community networks; all held a DVLA-issued licence and had at least one year's experience. Thirteen respondents spanning ages, backgrounds, and driving environments joined two audio-recorded, semi-structured focus groups, using a documented guide, to examine privacy, tone, and context through interaction. Interviews with UK drivers ($n=10$), police officers ($n=4$), and instructors ($n=2$) provided confidential individual and practitioner accounts. Survey responses were summarised descriptively. Qualitative recordings were transcribed verbatim and analysed inductively; the primary researcher iteratively grouped interview codes into higher-level themes. Triangulation across structured breadth, group interaction, and individual depth yielded requirements for content, timing, context, tone, and privacy \citep{sourceStudyA2026}.

The UK-focused evidence emphasised turning numerical events into explanations: what occurred, the relevant circumstances, why it was risky, and what to do next. Some participants also wanted educational reminders about legal thresholds, fines, and penalty points, but legal instruction was one requirement rather than the principal organising function of the reports. Participants considered real-time alerts potentially distracting and preferred daily or weekly post-trip summaries. Harsh or judgemental language was expected to reduce engagement; neutral, practical and educational wording was favoured \citep{sourceStudyA2026}.

Place changed the meaning of an event. Interviews highlighted schools, medical facilities, shopping areas, and other places with vulnerable road users. Consequently, speeding near a school was not merely another speed-limit breach: the report could connect it to children and time-specific local risk. Road type also shaped advice, including following distance on motorways, overtaking and unexpected hazards on rural roads, and crowded places in urban areas. Legal reminders remained a possible requirement, particularly when rules or penalties change, but they were not the principal content driver.

\subsection{Nigeria: education, authority, and alcohol risk}

To investigate Nigeria's local driving context, safety priorities, and feedback requirements, Safe Drive Africa conducted a mixed-methods stakeholder requirements study \citep{sourceStudyB2025}. A mixed-methods design was chosen to capture both broad statistical patterns of driving behaviour and rich qualitative explanations of local road challenges. Stratified random sampling was used to recruit 80 key stakeholders across two complementary groups: 51 road drivers (commercial, private, and fleet) and 29 officials from the Federal Road Safety Corps (FRSC), Nigeria's federal road-safety regulator and enforcement agency. Data were collected over two months through face-to-face questionnaire distribution and a secure online form. Role-tailored parallel questionnaires contained 25 questions for drivers and 11 for FRSC officials, combining demographic, binary, multiple-choice, and open-ended items. For analysis, open-ended responses underwent inductive thematic analysis in NVivo to uncover key risk rationales, while quantitative survey items were analysed using descriptive statistics and Fisher's exact tests \citep{sourceStudyB2025}.

The thematic analysis identified four principal unsafe-driving themes: non-compliant driving, aggressive driving, distracted driving, and driving in fatigue. Participants attributed these behaviours to driver habits, including alcohol/substance use and poor vehicle maintenance, limited formal training and traffic-rule knowledge, and external factors such as poor road conditions and traffic intensity \citep{sourceStudyB2025}. These findings are consistent with Nigerian evidence on road-user attitudes, road-safety constraints, and locally salient speeding beliefs \citep{uhegbu2021attitudes,uzondu2022roadsafety,etika2021beliefs}. Crucially for content design, ``lack of training'' covered limited formal instruction, limited traffic-rule knowledge, and difficulty interpreting signs. Stakeholders called for continuing education and stricter enforcement; 78 of 80 (97.5\%) supported a monitoring-and-feedback app \citep{sourceStudyB2025}. This requirements-led approach also responds to reviews calling for road-safety mobile systems to cover the full information cycle rather than detection alone \citep{aghayari2023mobile}.

Those findings motivated two content functions. Event-triggered Tips address foundational knowledge by citing applicable Nigerian rules, fines, and penalties. Weekly Reports address habits and attitudes through supportive reflection. Alcohol-influenced driving received dedicated treatment because stakeholders repeatedly identified it as a major local risk and requested detection of alcohol consumption. The project subsequently collected locally labelled smartphone-sensor data and trained a post-trip alcohol classifier on 152 trips (reported nested-cross-validation area under the receiver operating characteristic curve (AUC) 0.855). Alcohol status can therefore be acknowledged empathetically in a Report, with plan-ahead alternatives when risk is detected and positive reinforcement otherwise.

\subsection{Comparative boundary}

The contrast does not imply that UK drivers know every rule, that all Nigerian drivers lack licences, or that alcohol-impaired driving is unique to Nigeria. Rather, the local studies supplied different evidence about the content that each intervention needed to foreground.

A separate exploratory strand provides further requirements evidence without extending the comparison to a third system. Fifteen semi-structured interviews with Pakistan-based Uber drivers, conducted in Urdu or English and analysed inductively, highlighted dense traffic, inconsistent signage, unpredictable pedestrian activity, weak enforcement, and a preference for daily or weekly reports with reminders linked to markets, schools, mosques, and busy junctions \citep{sourceStudyA2026}. No feedback system was implemented or evaluated in Pakistan, so these observations illustrate another potential context for adaptation rather than evidence of system effectiveness. The analysis is consequently limited to study-specific findings instead of attributing uniform characteristics to national populations.

%% file: sections/systems.tex
\section{Two Localised Content Strategies}

\subsection{DRIVINGBEACON: contextual post-trip reflection}

DRIVINGBEACON produces daily or weekly in-app narrative reports from UK trip summaries. Its content planner selects detected speeding, mobile-phone use, harsh braking and acceleration, trip time, road type, and proximity to geofenced sensitive locations. It then combines events into a readable account rather than presenting a dashboard score alone (see Figure~\ref{fig:screenshots}(\subref{fig:uk-report})) \citep{sourceStudyA2026}.

Three content properties follow from the requirements. First, explanations are \emph{situated}: a report can identify the event and connect it to a familiar location or road condition. Second, advice is \emph{actionable}: the report explains why the behaviour matters and suggests an achievable change. Third, wording is \emph{socially acceptable}: a lightweight profile records age group, predominant driving environment, sensitivity/stress level, preferred feedback style, and report frequency. These attributes condition tone and emphasis, not the measured facts. A user can receive friendly or direct language, but an event cannot disappear because of a profile.

This strategy treats feedback as supported reflection for drivers already operating in a comparatively formal licensing and enforcement setting. It can include brief legal information, but its distinctive value is turning telematics into a fair, contextually specific narrative. Pilot participants preferred enhanced reports containing phone-use and geofenced context to basic event summaries, although the eight-person, four-week pilot was not powered to show behavioural efficacy.

\subsection{Safe Drive Africa: education plus persuasion}

Safe Drive Africa produces two deliberately separate forms of feedback for Nigerian drivers \citep{sourceStudyB2025}. \textbf{Tips} (Figure~\ref{fig:screenshots}(\subref{fig:ng-tip})) are short messages generated and displayed once per day from unsafe behaviours detected that day, such as harsh braking, rapid acceleration, or swerving; they are not presented continuously while the driver is driving. Behaviour keywords retrieve entries from a curated knowledge base built from the Nigerian Highway Code and National Road Traffic Regulations. The prompt requires applicable law sections, penalties, and fine values to be copied from the retrieved context, while adding a plain-language explanation and practical action. Tips therefore identify the unsafe behaviour and provide the corresponding rule and consequence.

\textbf{Reports} (Figure~\ref{fig:screenshots}(\subref{fig:ng-report})) are weekly, driver-specific narratives over structured statistics: total and most frequent unsafe behaviours, an example date/time and road, trip duration and distance, locations, and alcohol status. A two-stage generation-and-reflection procedure applies TPB principles implicitly. It praises safe habits, acknowledges local realities such as congestion, poor roads, or rain, gives two or three ability-building actions, and uses a supportive tone. Alcohol-positive reports directly but empathetically recommend not driving after drinking and planning alternatives; alcohol-negative reports reinforce the safe pattern. The Report is therefore not a longer Tip: the former supports motivation and perceived control, while the latter supplies precise legal education.

The content inventory is bounded by measurement. The app uses smartphone accelerometer, gyroscope, and Global Positioning System (GPS) streams for motion and unsafe-event detection. Its post-trip alcohol model was trained and evaluated using voluntary daily alcohol self-reports as ground-truth labels; those self-reports are not operational input features to the classifier. Fatigue and phone distraction were raised by stakeholders but omitted because they could not be measured credibly without extra hardware. Speeding was also omitted from the principal evaluation measure because reliable digital speed-limit metadata were not consistently obtainable.

\begin{table}[t]
\small
\centering
\begin{tabularx}{\columnwidth}{@{}>{\raggedright\arraybackslash}p{0.23\columnwidth}>{\raggedright\arraybackslash}X@{}}
\toprule
Content layer & Locally motivated configuration \\
\midrule
Core purpose & \textbf{UK:} explain post-trip events in context.\newline \textbf{Nigeria:} teach rules and support behaviour change. \\
Evidence & \textbf{UK:} event, time, and road/place.\newline \textbf{Nigeria:} event statistics plus retrieved law, fine, and penalty. \\
Risk focus & \textbf{UK:} speed, phone use, manoeuvres, and sensitive places.\newline \textbf{Nigeria:} manoeuvres and alcohol influence. \\
Adaptation & \textbf{UK:} profile-sensitive tone and road advice.\newline \textbf{Nigeria:} legal retrieval, local realities, and TPB coaching. \\
Surface & \textbf{UK:} integrated daily/weekly Report.\newline \textbf{Nigeria:} event Tips plus weekly Report. \\
\bottomrule
\end{tabularx}
\caption{The systems implement different locally motivated content strategies.}
\label{tab:content}
\end{table}

%% file: sections/architecture.tex
\section{How the Pipelines Support the Content}

Architecture is secondary to the content argument, but it explains how each strategy is made dependable (Table~\ref{tab:architecture}). Both systems separate sensed data from user-facing language and use a large language model (LLM) only after structured evidence has been assembled. They differ in where content authority resides.

\begin{table*}[t]
\small
\centering
\begin{tabularx}{\textwidth}{@{}>{\raggedright\arraybackslash}p{0.08\textwidth}>{\raggedright\arraybackslash}X>{\raggedright\arraybackslash}X>{\raggedright\arraybackslash}X>{\raggedright\arraybackslash}X@{}}
\toprule
System & Local evidence & Content selection & Generation control & Driver-facing output \\
\midrule
UK & Telematics events, speed-limit fields, maps, and geofences & Rules select incidents and road/place advice & Constrained paraphrase of the rule-selected draft & Contextual daily or weekly Report \\
Nigeria & Smartphone events, alcohol classification, and legal knowledge base & Keyword legal retrieval for Tips; weekly statistics for Reports & Structured Tip output; two-step Report generation and revision & Once-daily legal Tips; weekly persuasive Reports \\
\bottomrule
\end{tabularx}
\caption{How locally available evidence determines content selection, generation control, and feedback delivery in the UK and Nigerian systems.}
\label{tab:architecture}
\end{table*}

\paragraph{Factual control in DRIVINGBEACON.}
Cloud telematics supplies trip and waypoint data, including measured speed and available speed limits; mapping and geofences add road and place context. Deterministic rules decide which facts and advice enter the draft. The LLM is limited to paraphrasing the resulting report under instructions to preserve numbers and behavioural meaning; it does not select incidents or infer new facts. This yields an audit path from each report statement back to a trip variable or rule while still permitting tone variation.

\paragraph{Dual grounding in Safe Drive Africa.}
Deployment context also influenced component placement. DRIVINGBEACON could rely on cloud telematics in its UK deployment, whereas intermittent mobile connectivity and data cost in Nigeria motivated local persistence and on-device processing in Safe Drive Africa. The content-relevant distinction is between two evidence packages. A Tip is grounded by the detected event and retrieved Nigerian regulation. Its prompt prohibits invented laws, fines, or penalties and returns structured fields. A Report is grounded by aggregated driver statistics and passes through a second reflective prompt that checks coherence and implicit TPB coverage. Thus legal accuracy and motivational narrative use different controls instead of forcing both into one prompt.

\paragraph{Availability of speed-limit information.}
In the UK deployment, the Damoov telematics service supplied waypoint-level vehicle-speed and corresponding speed-limit fields, allowing evidence-based speeding feedback. No equivalent sufficiently reliable digital source of posted speed limits was available for the Nigerian deployment. The source study reports that roadside speed-limit signs were missing, inconsistent, or undocumented \citep{sourceStudyB2025}. GPS speed alone therefore could not establish a legal breach reliably. The Nigerian system excluded speeding from its unsafe-events-per-kilometre measure rather than generating an unsupported claim.

\paragraph{Models and inference settings.}\mbox{}\par
\noindent\textbf{UK system:} DRIVINGBEACON used OpenAI's Chat Completions application programming interface (API) with \texttt{gpt-3.5-turbo}, sending the assembled evidence and profile instructions as one user-role message. Its source study does not report decoding settings for this call.

\noindent\textbf{Nigerian system:} Safe Drive Africa sent the same retrieved legal context independently to \texttt{GPT-5.1} and \texttt{Gemini Nano Banana Pro} for Tips and displayed both outputs. Its OpenAI Tip call used temperature 0.5 and \texttt{max\_tokens=750}; its Gemini Tip call used Google Generative AI Android Software Development Kit (SDK) 0.9.0 with no explicitly configured decoding parameters. Reports used \texttt{gpt-4-turbo} for initial generation and \texttt{gpt-4o-mini} for mandatory revision, both with temperature 0.0 and \texttt{max\_tokens=350}. Safe Drive Africa's OpenAI requests used the Chat Completions endpoint, \url{https://api.openai.com/v1/chat/completions}. For its generation calls, top-$p$, seed, safety settings, and dated model snapshots were not specified, so provider or SDK defaults applied.

\begin{figure*}[t]
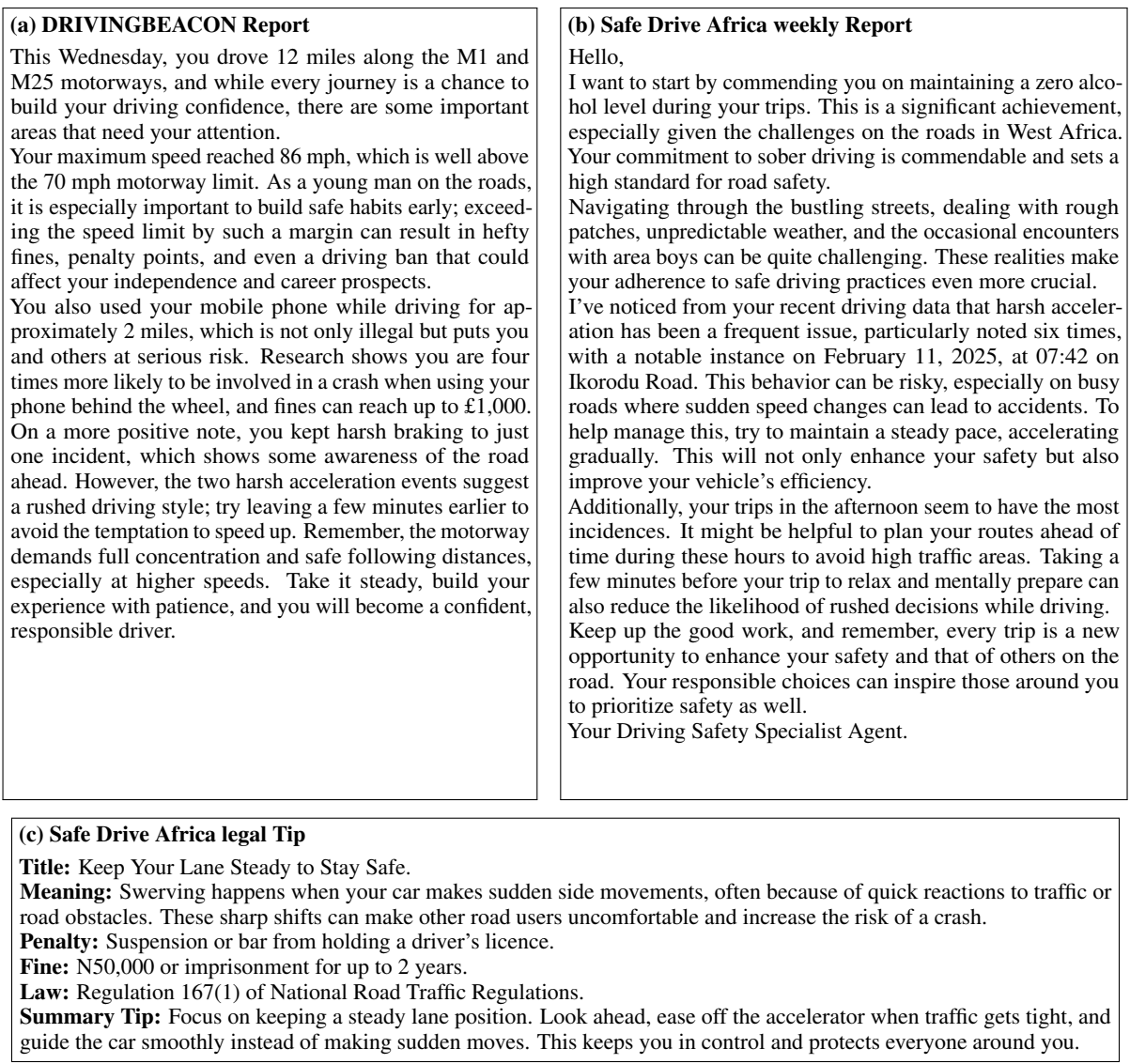

\centering
\begin{subfigure}[t]{0.48\textwidth}
\centering
\phantomcaption\label{fig:uk-report}
\fbox{\begin{minipage}[t][4.3in][t]{0.95\linewidth}\small
\textbf{(a) DRIVINGBEACON Report}\par\smallskip
This Wednesday, you drove 12 miles along the M1 and M25 motorways, and while every journey is a chance to build your driving confidence, there are some important areas that need your attention.

Your maximum speed reached 86 mph, which is well above the 70 mph motorway limit. As a young man on the roads, it is especially important to build safe habits early; exceeding the speed limit by such a margin can result in hefty fines, penalty points, and even a driving ban that could affect your independence and career prospects.

You also used your mobile phone while driving for approximately 2 miles, which is not only illegal but puts you and others at serious risk. Research shows you are four times more likely to be involved in a crash when using your phone behind the wheel, and fines can reach up to \pounds1,000.

On a more positive note, you kept harsh braking to just one incident, which shows some awareness of the road ahead. However, the two harsh acceleration events suggest a rushed driving style; try leaving a few minutes earlier to avoid the temptation to speed up. Remember, the motorway demands full concentration and safe following distances, especially at higher speeds. Take it steady, build your experience with patience, and you will become a confident, responsible driver.
\end{minipage}}
\end{subfigure}\hfill
\begin{subfigure}[t]{0.51\textwidth}
\centering
\phantomcaption\label{fig:ng-report}
\fbox{\begin{minipage}[t][4.3in][t]{0.95\linewidth}\small
\textbf{(b) Safe Drive Africa weekly Report}\par\smallskip
Hello,

I want to start by commending you on maintaining a zero alcohol level during your trips. This is a significant achievement, especially given the challenges on the roads in West Africa. Your commitment to sober driving is commendable and sets a high standard for road safety.

Navigating through the bustling streets, dealing with rough patches, unpredictable weather, and the occasional encounters with area boys can be quite challenging. These realities make your adherence to safe driving practices even more crucial.

I've noticed from your recent driving data that harsh acceleration has been a frequent issue, particularly noted six times, with a notable instance on February 11, 2025, at 07:42 on Ikorodu Road. This behavior can be risky, especially on busy roads where sudden speed changes can lead to accidents. To help manage this, try to maintain a steady pace, accelerating gradually. This will not only enhance your safety but also improve your vehicle's efficiency.

Additionally, your trips in the afternoon seem to have the most incidences. It might be helpful to plan your routes ahead of time during these hours to avoid high traffic areas. Taking a few minutes before your trip to relax and mentally prepare can also reduce the likelihood of rushed decisions while driving.

Keep up the good work, and remember, every trip is a new opportunity to enhance your safety and that of others on the road. Your responsible choices can inspire those around you to prioritize safety as well.

Your Driving Safety Specialist Agent.
\end{minipage}}
\end{subfigure}\par\medskip
\begin{subfigure}[t]{0.98\textwidth}
\centering
\phantomcaption\label{fig:ng-tip}
\fbox{\begin{minipage}[t]{0.98\linewidth}\small
\textbf{(c) Safe Drive Africa legal Tip}\par\smallskip
\textbf{Title:} Keep Your Lane Steady to Stay Safe.\par
\textbf{Meaning:} Swerving happens when your car makes sudden side movements, often because of quick reactions to traffic or road obstacles. These sharp shifts can make other road users uncomfortable and increase the risk of a crash.\par
\textbf{Penalty:} Suspension or bar from holding a driver's licence.\par
\textbf{Fine:} N50,000 or imprisonment for up to 2 years.\par
\textbf{Law:} Regulation 167(1) of National Road Traffic Regulations.\par
\textbf{Summary Tip:} Focus on keeping a steady lane position. Look ahead, ease off the accelerator when traffic gets tight, and guide the car smoothly instead of making sudden moves. This keeps you in control and protects everyone around you.
\end{minipage}}
\end{subfigure}
\caption{Source-study examples: (a) a DRIVINGBEACON Report; (b) a Safe Drive Africa weekly Report; and (c) a Safe Drive Africa legal Tip.}
\label{fig:screenshots}
\end{figure*}

\paragraph{Illustrative feedback surfaces.}
Figure~\ref{fig:screenshots}(\subref{fig:uk-report}) shows how the UK surface integrates incidents and contextual reflection in a report. The Nigerian surfaces separate a longer weekly coaching narrative (Figure~\ref{fig:screenshots}(\subref{fig:ng-report})) from a legally grounded Tip (Figure~\ref{fig:screenshots}(\subref{fig:ng-tip})).

%% file: sections/evaluation.tex
\section{Evidence of Effectiveness}

The two field evaluations examine whether locally designed feedback is associated with safer behaviour, but they do not provide a head-to-head test. Table~\ref{tab:evaluation} therefore reports within-study evidence and makes non-equivalence explicit.

\begin{table*}[t]
\small
\centering
\begin{tabularx}{\textwidth}{@{}>{\raggedright\arraybackslash}p{0.12\textwidth}>{\raggedright\arraybackslash}X>{\raggedright\arraybackslash}X@{}}
\toprule
Dimension & DRIVINGBEACON: UK & Safe Drive Africa: Nigeria \\
\midrule
Design & 12-week within-subject deployment: 4-week pre-intervention, 5-week intervention, and 3-week post-intervention with feedback withdrawn. & Within-subject pre-intervention/intervention deployment over approximately 25 weeks, excluding transition trips. \\
Data & 30 drivers; daily or weekly Reports; phase-level driver event rates. & 54 analysed drivers (27 commercial, 14 private, 13 company); 802 trips; once-daily Tips and weekly Reports. \\
Outcome & Unsafe events/km from speeding, phone use, and harsh braking/acceleration; geofenced events remain in the total and place context supports interpretation. & Unsafe events/km from harsh braking, rapid acceleration, and swerving; speeding excluded. \\
Result & 0.4275 pre-intervention, 0.2270 intervention, and 0.1493 post-intervention; significant phase effect and all pairwise differences. & 1.44 pre-intervention and 1.16 intervention, a 19.4\% reduction ($p<.001$); 43/54 drivers (79.6\%) improved. \\
Limitations & No parallel control group; all drivers received feedback during intervention; the personalisation effect was not isolated; post-intervention observation lasted three weeks. & No parallel control group; intervention trip count did not correlate significantly with improvement; alcohol-classifier outputs were excluded from the outcome. \\
\bottomrule
\end{tabularx}
\caption{Evaluation summary. Rates use a common distance unit, but different event definitions, durations, and designs make effect sizes non-comparable.}
\label{tab:evaluation}
\end{table*}

\subsection{UK deployment}

The 30-driver study recorded four weeks without reports (pre-intervention), five weeks with personalised contextual reports (intervention), and three weeks after reports were withdrawn (post-intervention). \citet{sourceStudyA2026} reports means of 0.6881, 0.3654, and 0.2403 incidents per mile; dividing by 1.609344 gives \emph{unsafe events per kilometre}: 0.4275 pre-intervention, 0.2270 intervention, and 0.1493 post-intervention. Lower rates indicate fewer detected unsafe events: these values correspond to approximately 43, 23, and 15 events per 100 kilometres. The intervention mean was 46.9\% below the pre-intervention mean, while the post-intervention mean was a further 34.2\% lower, or 65.1\% below pre-intervention. A repeated-measures one-way analysis of variance (ANOVA) on the original per-mile rates found a significant phase effect, $F(2,58)=85.2$, $p<.0001$. Tukey honestly significant difference comparisons were significant for pre-intervention versus intervention ($p<.0001$), intervention versus post-intervention ($p=.0018$), and pre-intervention versus post-intervention ($p<.0001$). This pattern is consistent with improvement during the intervention and short-term maintenance, but the three-week post-intervention phase is too short to establish durable behaviour change. Because all participants received the complete intervention, the study cannot isolate tone personalisation, contextual content, or observation effects.

Qualitative responses illuminate content reception. Participants described the reports as helpful and non-blaming, valued friendly or profile-appropriate tone, and reported that references to speeding near schools made an abstract event feel consequential. Suggested improvements included visual or map summaries, adjustable report frequency and detail, audio summaries, and less repetitive wording. These observations support acceptability of the UK strategy, though they are not independent evidence of efficacy.

\subsection{Nigerian deployment}

The Nigerian study analysed 54 drivers and 802 trips. Driver-level unsafe behaviours per kilometre (UBPK) fell from 1.44 to 1.16, a 19.4\% mean reduction; 43 drivers (79.6\%) improved \citep{sourceStudyB2025}. A Wilcoxon signed-rank test found the pre/intervention difference significant ($p<.001$). The study also introduced \emph{bad-day frequency}: the proportion of intervention trips whose trip-level UBPK exceeded the 75th percentile of that driver's pre-intervention trip-level UBPK distribution. This driver-specific threshold represents the riskiest quarter of pre-intervention trips. Drivers reaching similar final event rates sometimes followed very different trajectories, from steady improvement with no bad days to volatile improvement with repeated relapses. This suggests an NLG opportunity beyond changing average risk: content cadence and emphasis could respond to stability.

The ancillary post-trip alcohol classifier was a bagged decision-tree model evaluated on 152 labelled trips using nested $5\times5$ cross-validation. Beyond AUC 0.855 (95\% confidence interval (CI) 0.794--0.909), it achieved mean F1 0.589 (95\% CI 0.435--0.711) and mean recall 0.615 (95\% CI 0.433--0.773). Aggregated out-of-fold predictions yielded accuracy 0.829, precision 0.576, recall 0.613, F1 0.594, and specificity 0.884. These results describe trip classification, not the behavioural-intervention outcome \citep{sourceStudyB2025}.

There was no significant correlation between the number of intervention trips and the reduction in unsafe events per kilometre (Pearson $r=-0.071$, $p=.610$; Spearman $\rho=-0.043$, $p=.757$). The evidence therefore does not support a simple ``more app use causes more improvement'' claim. Four follow-up interviews instead help explain perceived usefulness: participants valued legal references and fines as credible, locally relevant support, while supportive wording reduced the sense of punishment. This aligns with the dual content rationale but remains a small qualitative sample.

\subsection{Scope of the comparison}

Both studies show statistically supported reductions in unsafe events per kilometre while locally adapted feedback was deployed, and both report favourable reception of the content features that distinguish them. Neither includes a contemporaneous randomised control, so ``associated with'' is more defensible than ``caused''. The common unit does not make the rates equivalent: the two measures contain different event sets, most notably because speeding is excluded in Nigeria. Ranking the systems by rate or percentage change would therefore be invalid. The comparative result is instead a replication at the design-principle level: locally derived content can be made operational and evaluated in two materially different settings.

%% file: sections/discussion.tex
\section{A Design Process for Localised NLG}

The comparison yields a five-stage design process, consistent with person-based and mobile road-safety design guidance \citep{yardley2015person,aghayari2023mobile}: (1) \emph{requirements elicitation} identifies stakeholder needs, risks, languages, and feedback timing; (2) \emph{evidence audit} determines which behaviours, contextual variables, and authorities are available and credible; (3) \emph{message definition} maps input data and evidence to information that the system may communicate; (4) \emph{message enrichment} adds supported information such as legal consequences, practical advice, or encouragement; and (5) \emph{report generation} realises the selected messages and additional information in the language and form required by stakeholders. Following standard NLG terminology, a message is a domain-level unit of information selected for communication before its wording is chosen \citep{reiter2000building}. This process is derived from two deployments, but crossover or component-level experiments are needed to establish which choices are genuinely locale-specific.

\paragraph{Stage 1: Requirements elicitation.}
In both systems, localisation began with eliciting local stakeholder needs, prevalent risks, and existing knowledge rather than adopting default assumptions. DRIVINGBEACON's interviews revealed that UK drivers wanted contextual post-trip reflection, sensitivity to personal feedback styles, and explanations tied to sensitive places like schools. Safe Drive Africa's stakeholder survey identified gaps in formal training and traffic-rule knowledge, alongside alcohol-influenced driving, creating a clear demand for educational rules, authoritative legal backing, and alcohol-aware coaching. Importantly, requirements elicitation must characterize the specific study population and deployment context without generating sweeping national stereotypes: the findings reflect elicited needs from the respective cohorts rather than uniform traits of all UK or Nigerian drivers.

\paragraph{Stage 2: Evidence audit.}
A safety-critical NLG system must systematically audit local data infrastructure to determine which behaviours, contextual variables, and authorities can be verified reliably. DRIVINGBEACON could discuss speeding because cloud telematics supplied waypoint-level vehicle speed and reliable posted speed limits. Conversely, because digital speed-limit metadata were sparse and unverified in the Nigerian deployment, Safe Drive Africa excluded speeding from both driver feedback and its evaluation metric. The core principle of evidence auditing is strict: when a required contextual variable cannot be grounded in local data, the system should omit the claim entirely rather than infer or hallucinate ungrounded assertions.

\paragraph{Stage 3: Message definition.}
Message definition maps audited sensor observations to the core communicative units that the system intends to convey. Rather than deploying a universal inventory of driving events, each system defined messages reflecting its audited risks: DRIVINGBEACON focused on speeding, phone distraction, and manoeuvring near sensitive places, whereas Safe Drive Africa defined messages around harsh manoeuvres, trip timing, and classifier-detected alcohol status. In both systems, a message represents a structured domain-level entity established before lexical realisation.

\paragraph{Stage 4: Message enrichment.}
Once core messages are defined, message enrichment attaches supported domain knowledge to make the feedback persuasive, actionable, and socially acceptable. In DRIVINGBEACON, enrichment incorporated driver profile attributes (age group, feedback sensitivity) to condition narrative tone, alongside place-specific risk information (such as risks to pedestrians near schools). In Safe Drive Africa, enrichment retrieved applicable laws, regulations, and statutory penalties via a curated legal knowledge base for daily Tips, and structured motivational advice using TPB principles (acknowledging local road conditions, offering ability-building tips, and praising sober trips) for weekly Reports.

\paragraph{Stage 5: Report generation and delivery.}
The final stage realises the enriched messages into user-facing text, selecting generation controls and delivery cadences matched to content function. DRIVINGBEACON generates integrated daily or weekly Reports using rule-based selection paired with constrained LLM paraphrasing to maintain factual integrity. Safe Drive Africa employs dual generation pathways: deterministic legal retrieval with structured LLM constraints for once-daily Tips, and a two-stage LLM generation-and-reflection pipeline for weekly persuasive Reports. Both deployments delivered feedback in English. For deployments in multilingual environments, report generation would also require establishing stakeholder language requirements and validating outputs in target languages rather than treating machine translation as sufficient localisation.

\section{Conclusion}

The comparison is not primarily between two software stacks, but between two locally motivated content strategies. UK feedback prioritises contextual post-trip reflection and tone-sensitive explanation. Nigerian feedback places greater emphasis on legal and safety education, locally identified alcohol risk, and persuasive weekly coaching. Data availability further determines the scope of evidentially supported content: UK speed-limit context supports speeding narratives, while Nigerian metadata scarcity requires omission of that claim and metric. Both deployments provide qualified evidence of behavioural improvement in context. The findings connect local evidence to content choice, grounding, generation control, and evaluation, without treating either country's system as a universal default.

\paragraph{Data availability.}
\citet{sourceStudyA2026} reports aggregate results and an anonymised phase-level extract for 10 drivers. Participant-level UK data are not publicly released because of privacy constraints. Safe Drive Africa participant-level and trip-level data are likewise not publicly released because location traces, driving-behaviour records, and alcohol self-reports create privacy constraints; aggregate results are reported in the source study \citep{sourceStudyB2025}.

\section{Limitations}

This is a comparative synthesis of two source studies, not a preregistered cross-country experiment. The systems differ in recruitment, duration, sensors, feedback exposure, event definitions, and statistical analysis. The requirements samples cannot represent all drivers in either country, and the four Nigerian follow-up interviews particularly limit claims about reception. The alcohol classifier predicts trip-level patterns associated with self-reported consumption; it is not a breathalyser or legal determination of impairment. The paper does not independently evaluate generated-text factuality, cultural resonance, or whether each content strategy would outperform the other in the same country.

\section{Ethical Considerations}

Driver monitoring raises privacy, surveillance, and fairness concerns. Both projects obtained institutional ethical approval and informed consent. Location traces, behavioural profiles, and alcohol self-reports are sensitive; deployment beyond research would require data minimisation, transparent access and retention policies, meaningful opt-out, and protection against punitive secondary use, consistent with empirical privacy preferences and regulatory analyses of automotive telematics \citep{dowthwaite2024privacy,truby2024regulatory}. Legal content also requires versioning and review when regulations change. The analysis consequently reports study-specific evidence and uncertainty without associating observed knowledge or infrastructure gaps with nationality itself.

%% file: refs.bib
@phdthesis{sourceStudyA2026,
  author = {Jawwad Baig},
  title = {Personalised Data-to-Text Feedback for Safer Driving},
  school = {University of Aberdeen},
  year = {2026},
  note = {Doctoral thesis manuscript}
}

@phdthesis{sourceStudyB2025,
  author = {Iniakpokeikiye Peter Thompson},
  title = {Using an {AI} App to Encourage Safer Driving in Non-Western Countries: A Nigerian Context},
  school = {University of Aberdeen},
  year = {2025}
}

@inproceedings{sourcePubA2022,
  author = {Jawwad Baig and Guanyi Chen and Chenghua Lin and Ehud Reiter},
  title = {{DrivingBeacon}: Driving Behaviour Change Support System Considering Mobile Use and Geo-information},
  booktitle = {Proceedings of the First Workshop on Natural Language Generation in Healthcare},
  pages = {1--8},
  publisher = {Association for Computational Linguistics},
  year = {2022},
  url = {https://aclanthology.org/2022.nlg4health-1.1/}
}

@article{sourcePubB2025,
  author = {Iniakpokeikiye Peter Thompson and Dewei Yi and Ehud Reiter},
  title = {An End-to-End System for Culturally-Attuned Driving Feedback using a Dual-Component {NLG} Engine},
  journal = {arXiv preprint arXiv:2509.04478},
  year = {2025},
  doi = {10.48550/arXiv.2509.04478},
  url = {https://arxiv.org/abs/2509.04478}
}

@article{braun2018saferdrive,
  author = {Daniel Braun and Ehud Reiter and Advaith Siddharthan},
  title = {SaferDrive: An NLG-based behaviour change support system for drivers},
  journal = {Natural Language Engineering},
  volume = {24},
  number = {4},
  pages = {551--588},
  year = {2018},
  doi = {10.1017/S1351324918000050},
  url = {https://doi.org/10.1017/S1351324918000050}
}

@article{reiter2003smoking,
  author = {Ehud Reiter and Roma Robertson and Liesl M. Osman},
  title = {Lessons from a failure: Generating tailored smoking cessation letters},
  journal = {Artificial Intelligence},
  volume = {144},
  number = {1--2},
  pages = {41--58},
  year = {2003},
  doi = {10.1016/S0004-3702(02)00370-3},
  url = {https://doi.org/10.1016/S0004-3702(02)00370-3}
}

@article{williams2008skillsum,
  author = {Sandra Williams and Ehud Reiter},
  title = {Generating basic skills reports for low-skilled readers},
  journal = {Natural Language Engineering},
  volume = {14},
  number = {4},
  pages = {495--525},
  year = {2008},
  doi = {10.1017/S1351324908004725},
  url = {https://doi.org/10.1017/S1351324908004725}
}

@book{reiter2000building,
  author = {Ehud Reiter and Robert Dale},
  title = {Building Natural Language Generation Systems},
  publisher = {Cambridge University Press},
  year = {2000},
  doi = {10.1017/CBO9780511519857},
  url = {https://doi.org/10.1017/CBO9780511519857}
}

@article{ajzen1991theory,
  author = {Icek Ajzen},
  title = {The theory of planned behavior},
  journal = {Organizational Behavior and Human Decision Processes},
  volume = {50},
  number = {2},
  pages = {179--211},
  year = {1991},
  doi = {10.1016/0749-5978(91)90020-T},
  url = {https://doi.org/10.1016/0749-5978(91)90020-T}
}

@inproceedings{fogg2009behavior,
  author = {BJ Fogg},
  title = {A behavior model for persuasive design},
  booktitle = {Proceedings of the 4th International Conference on Persuasive Technology},
  pages = {1--7},
  publisher = {ACM},
  year = {2009},
  doi = {10.1145/1541948.1541999},
  url = {https://doi.org/10.1145/1541948.1541999}
}

@inproceedings{novikova2017need,
  author = {Jekaterina Novikova and Ond{\v{r}}ej Du{\v{s}}ek and Amanda Cercas Curry and Verena Rieser},
  title = {Why We Need New Evaluation Metrics for {NLG}},
  booktitle = {Proceedings of the 2017 Conference on Empirical Methods in Natural Language Processing},
  pages = {2241--2252},
  publisher = {Association for Computational Linguistics},
  year = {2017},
  doi = {10.18653/v1/D17-1238},
  url = {https://aclanthology.org/D17-1238/}
}

@inproceedings{vanderlee2019best,
  author = {Chris van der Lee and Albert Gatt and Emiel van Miltenburg and Sander Wubben and Emiel Krahmer},
  title = {Best practices for the human evaluation of automatically generated text},
  booktitle = {Proceedings of the 12th International Conference on Natural Language Generation},
  pages = {355--368},
  publisher = {Association for Computational Linguistics},
  year = {2019},
  doi = {10.18653/v1/W19-8643},
  url = {https://aclanthology.org/W19-8643/}
}

@article{croxford2025llm,
  author = {Emma Croxford and Yanjun Gao and Nicholas Pellegrino and Karen Wong and Graham Wills and Elliot First and Frank Liao and Cherodeep Goswami and Brian Patterson and Majid Afshar},
  title = {Current and future state of evaluation of large language models for medical summarization tasks},
  journal = {npj Health Systems},
  volume = {2},
  number = {1},
  pages = {6},
  year = {2025},
  doi = {10.1038/s44401-024-00011-2},
  url = {https://doi.org/10.1038/s44401-024-00011-2}
}

@inproceedings{quintero2023ubi,
  author = {Juan Quintero and Victoria Karaseva and Freya Gassmann and Zinaida Benenson},
  title = {User Acceptance Factors of Usage-Based Insurance},
  booktitle = {Advances in Intelligent Traffic and Transportation Systems},
  year = {2023},
  doi = {10.3233/ATDE230018},
  url = {https://doi.org/10.3233/ATDE230018}
}

@article{li2023driving,
  author = {Hong-Jie Li and Xing-Gang Luo and Zhong-Liang Zhang and Wei Jiang and Shen-Wei Huang},
  title = {Driving risk prevention in usage-based insurance services based on interpretable machine learning and telematics data},
  journal = {Decision Support Systems},
  volume = {172},
  pages = {113985},
  year = {2023},
  doi = {10.1016/j.dss.2023.113985},
  url = {https://doi.org/10.1016/j.dss.2023.113985}
}

@article{ebert2026ubi,
  author = {Jeffrey P. Ebert and Catherine C. McDonald and Ruiying A. Xiong and Dina Abdel-Rahman and Neda Khan and Maya Irving and Athena Lee and Arjun Patel and Sadie Friday and Brian Tefft and William J. Horrey and Subhash Aryal and Michael O. Harhay and M. Kit Delgado},
  title = {A national randomized field experiment testing the impact of simulated usage-based insurance programs on driver safety},
  journal = {Accident Analysis \& Prevention},
  volume = {225},
  pages = {108331},
  year = {2026},
  doi = {10.1016/j.aap.2025.108331},
  url = {https://doi.org/10.1016/j.aap.2025.108331}
}

@inproceedings{singh2024cultural,
  author = {Pushpdeep Singh and Mayur Patidar and Lovekesh Vig},
  title = {Translating Across Cultures: {LLM}s for Intralingual Cultural Adaptation},
  booktitle = {Proceedings of the 28th Conference on Computational Natural Language Learning},
  pages = {400--418},
  publisher = {Association for Computational Linguistics},
  year = {2024},
  doi = {10.18653/v1/2024.conll-1.30},
  url = {https://aclanthology.org/2024.conll-1.30/}
}

@article{liu2025cultural,
  author = {Chen Cecilia Liu and Iryna Gurevych and Anna Korhonen},
  title = {Culturally Aware and Adapted {NLP}: A Taxonomy and a Survey of the State of the Art},
  journal = {Transactions of the Association for Computational Linguistics},
  volume = {13},
  pages = {652--689},
  year = {2025},
  doi = {10.1162/tacl_a_00760},
  url = {https://aclanthology.org/2025.tacl-1.31/}
}

@inproceedings{braun2015creating,
  author = {Daniel Braun and Ehud Reiter and Advaith Siddharthan},
  title = {Creating Textual Driver Feedback from Telemetric Data},
  booktitle = {Proceedings of the 15th European Workshop on Natural Language Generation ({ENLG})},
  pages = {156--165},
  publisher = {Association for Computational Linguistics},
  year = {2015},
  doi = {10.18653/v1/W15-4726},
  url = {https://aclanthology.org/W15-4726/}
}

@inproceedings{riostorres2016overview,
  author = {Jackeline Rios-Torres and Andreas A. Malikopoulos},
  title = {An Overview of Driver Feedback Systems for Efficiency and Safety},
  booktitle = {2016 IEEE 19th International Conference on Intelligent Transportation Systems},
  pages = {667--674},
  publisher = {IEEE},
  year = {2016},
  doi = {10.1109/ITSC.2016.7795625},
  url = {https://doi.org/10.1109/ITSC.2016.7795625}
}

@article{michelaraki2021posttrip,
  author = {Eva Michelaraki and Christos Katrakazas and George Yannis and Ashleigh Filtness and Rachel Talbot and Graham Hancox and Fran Pilkington-Cheney and Kris Brijs and Veerle Ross and H{\'e}l{\`e}ne Dirix and An Neven and Roeland Paul and Tom Brijs and Petros Fortsakis and Eleni Konstantina Frantzola and Rodrigo Taveira},
  title = {Post-trip Safety Interventions: State-of-the-art, Challenges, and Practical Implications},
  journal = {Journal of Safety Research},
  volume = {77},
  pages = {67--85},
  year = {2021},
  doi = {10.1016/j.jsr.2021.02.005},
  url = {https://doi.org/10.1016/j.jsr.2021.02.005}
}

@article{boylan2024telematics,
  author = {James Boylan and Denny Meyer and Won Sun Chen},
  title = {A Systematic Review of the Use of In-Vehicle Telematics in Monitoring Driving Behaviours},
  journal = {Accident Analysis \& Prevention},
  volume = {199},
  pages = {107519},
  year = {2024},
  doi = {10.1016/j.aap.2024.107519},
  url = {https://doi.org/10.1016/j.aap.2024.107519}
}

@article{kontaxi2025feedback,
  author = {Armira Kontaxi and Apostolos Ziakopoulos and George Yannis},
  title = {Exploring the Impact of Driver Feedback on Safety: A Systematic Review of Studies in Real-World Driving Conditions},
  journal = {Transportation Research Part F: Traffic Psychology and Behaviour},
  volume = {114},
  pages = {118--140},
  year = {2025},
  doi = {10.1016/j.trf.2025.05.028},
  url = {https://doi.org/10.1016/j.trf.2025.05.028}
}

@article{stevenson2021telematics,
  author = {Mark Stevenson and Anthony Harris and Jasper S. Wijnands and Duncan Mortimer},
  title = {The Effect of Telematic Based Feedback and Financial Incentives on Driving Behaviour: A Randomised Trial},
  journal = {Accident Analysis \& Prevention},
  volume = {159},
  pages = {106278},
  year = {2021},
  doi = {10.1016/j.aap.2021.106278},
  url = {https://doi.org/10.1016/j.aap.2021.106278}
}

@article{meuleners2023personalized,
  author = {Lynn Meuleners and Michelle Fraser and Mark Stevenson and Paul Roberts},
  title = {Personalized Driving Safety: Using Telematics to Reduce Risky Driving Behavior among Young Drivers},
  journal = {Journal of Safety Research},
  volume = {86},
  pages = {164--173},
  year = {2023},
  doi = {10.1016/j.jsr.2023.05.007},
  url = {https://doi.org/10.1016/j.jsr.2023.05.007}
}

@article{vankov2023provisional,
  author = {Daniel Vankov and Ronald Schroeter and Andry Rakotonirainy},
  title = {Provisional Drivers Intend to Speed Less: The Positive Outcome for Young Drivers of a Safe-Driving App Randomised Trial},
  journal = {Transportation Research Interdisciplinary Perspectives},
  volume = {21},
  pages = {100877},
  year = {2023},
  doi = {10.1016/j.trip.2023.100877},
  url = {https://doi.org/10.1016/j.trip.2023.100877}
}

@article{warren2018backpocketdriver,
  author = {Ian Warren and Andrew Meads and Robyn Whittaker and Rosie Dobson and Shanthi Ameratunga},
  title = {Behavior Change for Youth Drivers: Design and Development of a Smartphone-Based App (BackPocketDriver)},
  journal = {JMIR Formative Research},
  volume = {2},
  number = {2},
  pages = {e25},
  year = {2018},
  doi = {10.2196/formative.9660},
  url = {https://formative.jmir.org/2018/2/e25}
}

@article{gao2022telematics,
  author = {Guangyuan Gao and Shengwang Meng and Mario V. W{\"u}thrich},
  title = {What Can We Learn from Telematics Car Driving Data: A Survey},
  journal = {Insurance: Mathematics and Economics},
  volume = {104},
  pages = {185--199},
  year = {2022},
  doi = {10.1016/j.insmatheco.2022.02.004},
  url = {https://doi.org/10.1016/j.insmatheco.2022.02.004}
}

@article{ghaffarpasand2022vehicle,
  author = {Omid Ghaffarpasand and Mark Burke and Louisa K. Osei and Helen Ursell and Sam Chapman and Francis D. Pope},
  title = {Vehicle Telematics for Safer, Cleaner and More Sustainable Urban Transport: A Review},
  journal = {Sustainability},
  volume = {14},
  number = {24},
  pages = {16386},
  year = {2022},
  doi = {10.3390/su142416386},
  url = {https://doi.org/10.3390/su142416386}
}

@article{picco2023monitoring,
  author = {Ang{\`e}le Picco and Arjan Stuiver and Joost de Winter and Dick de Waard},
  title = {The Use of Monitoring and Feedback Devices in Driving: An Assessment of Acceptability and Its Key Determinants},
  journal = {Transportation Research Part F: Traffic Psychology and Behaviour},
  volume = {92},
  pages = {1--14},
  year = {2023},
  doi = {10.1016/j.trf.2022.10.021},
  url = {https://doi.org/10.1016/j.trf.2022.10.021}
}

@article{dowthwaite2024privacy,
  author = {Anna Dowthwaite and Dave Cook and Anna L. Cox},
  title = {Privacy Preferences in Automotive Data Collection},
  journal = {Transportation Research Interdisciplinary Perspectives},
  volume = {24},
  pages = {101022},
  year = {2024},
  doi = {10.1016/j.trip.2024.101022},
  url = {https://doi.org/10.1016/j.trip.2024.101022}
}

@article{truby2024regulatory,
  author = {Jon Truby and Rafael Dean Brown and Imad Antoine Ibrahim},
  title = {Regulatory Options for Vehicle Telematics Devices: Balancing Driver Safety, Data Privacy and Data Security},
  journal = {International Review of Law, Computers \& Technology},
  volume = {38},
  number = {1},
  pages = {86--110},
  year = {2024},
  doi = {10.1080/13600869.2023.2242671},
  url = {https://doi.org/10.1080/13600869.2023.2242671}
}

@article{somoray2024tpb,
  author = {Klaire Somoray and Katherine M. White and Barry Watson and Ioni Lewis},
  title = {Predicting Risky Driving Behaviours Using the Theory of Planned Behaviour: A Meta-Analysis},
  journal = {Accident Analysis \& Prevention},
  volume = {208},
  pages = {107797},
  year = {2024},
  doi = {10.1016/j.aap.2024.107797},
  url = {https://doi.org/10.1016/j.aap.2024.107797}
}

@article{yardley2015person,
  author = {Lucy Yardley and Leanne Morrison and Katherine Bradbury and Ingrid Muller},
  title = {The Person-Based Approach to Intervention Development: Application to Digital Health-Related Behavior Change Interventions},
  journal = {Journal of Medical Internet Research},
  volume = {17},
  number = {1},
  pages = {e30},
  year = {2015},
  doi = {10.2196/jmir.4055},
  url = {https://doi.org/10.2196/jmir.4055}
}

@inproceedings{oyibo2018persuasive,
  author = {Kiemute Oyibo and Ifeoma Adaji and Rita Orji and Babatunde Olabenjo and Julita Vassileva},
  title = {Susceptibility to Persuasive Strategies: A Comparative Analysis of Nigerians vs. Canadians},
  booktitle = {Proceedings of the 26th Conference on User Modeling, Adaptation and Personalization},
  pages = {229--238},
  publisher = {ACM},
  year = {2018},
  doi = {10.1145/3209219.3209239},
  url = {https://doi.org/10.1145/3209219.3209239}
}

@article{arslan2024culture,
  author = {Burcu Arslan and T{\"u}rker {\"O}zkan},
  title = {Role of Culture, Income Level and Governance Quality on Driver Behaviours},
  journal = {Journal of Road Safety},
  volume = {35},
  number = {3},
  pages = {10--23},
  year = {2024},
  doi = {10.33492/jrs-d-24-3-2319349},
  url = {https://doi.org/10.33492/jrs-d-24-3-2319349}
}

@article{uhegbu2021attitudes,
  author = {Uchenna Nnabuihe Uhegbu and Miles R. Tight},
  title = {Road User Attitudes and Their Reported Behaviours in Abuja, Nigeria},
  journal = {Sustainability},
  volume = {13},
  number = {8},
  pages = {4222},
  year = {2021},
  doi = {10.3390/su13084222},
  url = {https://doi.org/10.3390/su13084222}
}

@article{uzondu2022roadsafety,
  author = {Chinebuli Uzondu and Samantha Jamson and Greg Marsden},
  title = {Road Safety in Nigeria: Unravelling the Challenges, Measures, and Strategies for Improvement},
  journal = {International Journal of Injury Control and Safety Promotion},
  volume = {29},
  number = {4},
  pages = {522--532},
  year = {2022},
  doi = {10.1080/17457300.2022.2087230},
  url = {https://doi.org/10.1080/17457300.2022.2087230}
}

@article{etika2021beliefs,
  author = {Anderson Etika and Natasha Merat and Oliver Carsten},
  title = {Identifying Salient Beliefs Underlying Speeding Behaviour: An Elicitation Study of Nigerian Drivers},
  journal = {Transportation Research Interdisciplinary Perspectives},
  volume = {9},
  pages = {100279},
  year = {2021},
  doi = {10.1016/j.trip.2020.100279},
  url = {https://doi.org/10.1016/j.trip.2020.100279}
}

@article{aghayari2023mobile,
  author = {Hossein Aghayari and Leila R. Kalankesh and Homayoun Sadeghi-Bazargani and Mohammad-Reza Feizi-Derakhshi},
  title = {Functionalities of Mobile Solutions for Preventing Road Traffic Health and Safety Issues in the Context of Information Management Cycle},
  journal = {Transportation Research Record: Journal of the Transportation Research Board},
  volume = {2678},
  number = {4},
  pages = {649--658},
  year = {2024},
  doi = {10.1177/03611981231185769},
  url = {https://doi.org/10.1177/03611981231185769}
}

@article{osuji2024datatotext,
  author = {Chinonso Cynthia Osuji and Thiago Castro Ferreira and Brian Davis},
  title = {A Systematic Review of Data-to-Text {NLG}},
  journal = {arXiv preprint arXiv:2402.08496},
  year = {2024},
  doi = {10.48550/arXiv.2402.08496},
  url = {https://arxiv.org/abs/2402.08496}
}

@article{sharma2024neural,
  author = {Mandar Sharma and Ajay Kumar Gogineni and Naren Ramakrishnan},
  title = {Neural Methods for Data-to-Text Generation},
  journal = {ACM Transactions on Intelligent Systems and Technology},
  volume = {15},
  number = {5},
  pages = {1--46},
  year = {2024},
  doi = {10.1145/3660639},
  url = {https://doi.org/10.1145/3660639}
}
